\documentclass[10pt,twocolumn,letterpaper]{article}

\usepackage{wacv}
\usepackage{times}
\usepackage{epsfig}
\usepackage{graphicx}
\usepackage{amsmath}
\usepackage{amssymb}

\wacvfinalcopy % *** Uncomment this line for the final submission

\def\wacvPaperID{0188} % *** Enter the wacv Paper ID here

\ifwacvfinal\fi
\begin{document}

%%%%%%%%% TITLE
\title{One-to-one Mapping for Unpaired Image-to-image Translation}

% Authors at the same institution
\author{Zengming Shen \hspace{2cm} Yifan Chen\hspace{2cm} Thomas S. Huang \\
University of Illinois at Urbana-Champaign\\
{\tt\small zshen5,yifanc3,t-huang1@illinois.edu}
\and
S.Kevin Zhou \\
Institute of Computing Technology, Chinese Academy of Sciences\\
{\tt\small zhoushaohua@ict.ac.cn}
\and
Bogdan Georgescu \\
Siemens Healthineers\\
{\tt\small bogdan.georgescu@siemens-healthineers.com}
\and
Xuqi Liu \\
Rutgers University\\
{\tt\small xl325@scarletmail.rutgers.edu}
}
% Authors at different institutions
% \author{First Author \\
% \and
% Second Author \\

\maketitle
\ifwacvfinal\thispagestyle{empty}\fi

%%%%%%%%% ABSTRACT
\begin{abstract}
Recently image-to-image translation has attracted significant interests in the literature, starting from the successful use of the generative adversarial network (GAN), to the introduction of cyclic constraint, to extensions to multiple domains. However, in existing approaches, there is no guarantee that the mapping between two image domains is unique or one-to-one. Here we propose a self-inverse network learning approach for unpaired image-to-image translation. Building on top of CycleGAN, we learn a self-inverse function by simply augmenting the training samples by swapping inputs and outputs during training and with separated cycle consistency loss for each mapping direction. The outcome of such learning is a proven one-to-one mapping function. Our extensive experiments on a variety of datasets, including cross-modal medical image synthesis, object transfiguration, and semantic labeling, consistently demonstrate clear improvement over the CycleGAN method both qualitatively and quantitatively. Especially our proposed method reaches the state-of-the-art result on the cityscapes benchmark dataset for the label to photo unpaired directional image translation.
\end{abstract}

\section{Introduction}
Image-to-image translation (or cross-domain image synthesis) learns a mapping function from an input image to an output image or vice versa. It can grouped into two categories: supervised \cite{isola2017image} vs unsupervised (or unpaired) \cite{zhu2017unpaired}. 

\begin{figure}[t]
\begin{center}
  \includegraphics[width=\linewidth]{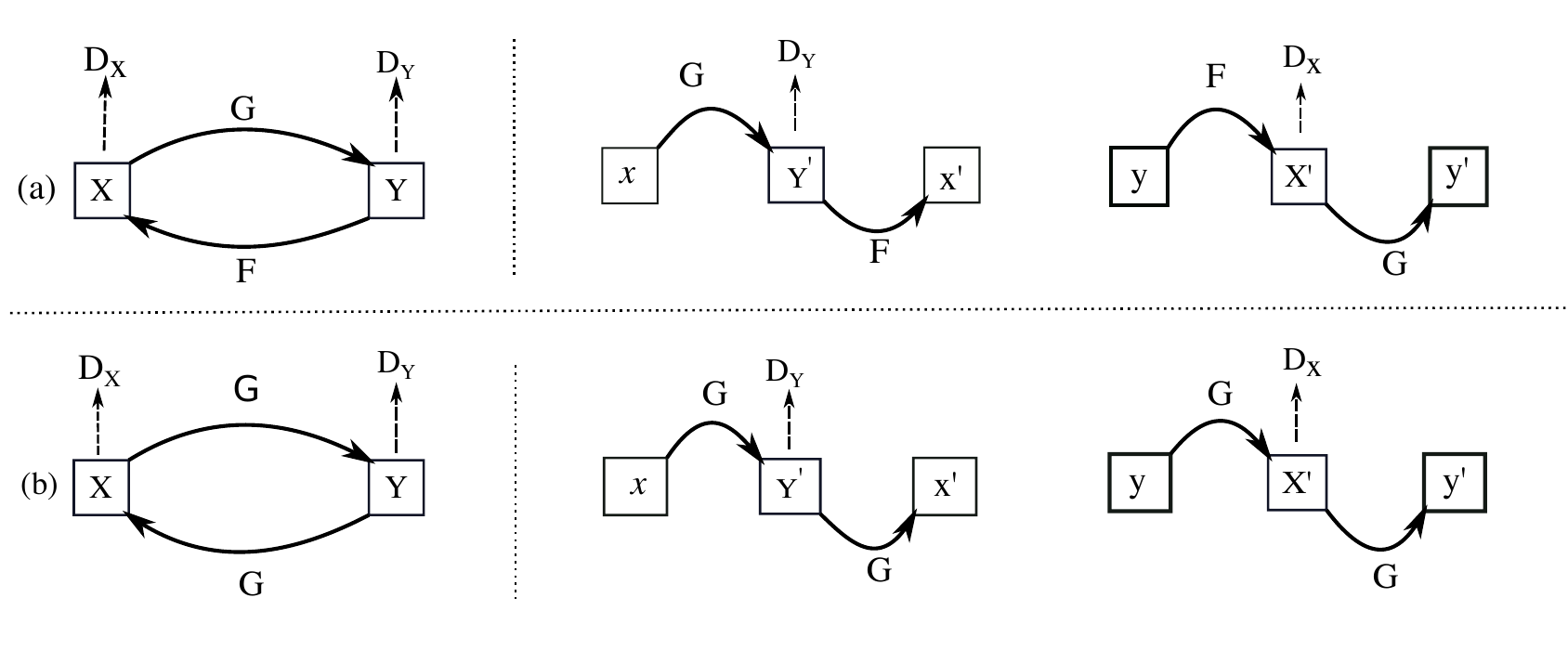}
\end{center}
   \caption{A comparison of our one2one CycleGAN with the original CycleGAN \cite{CycleGAN2017} for the mapping between two domains X and Y. (a) Original CycleGAN model. It contains two separated mapping functions $G: X \rightarrow Y$ and $F: Y \rightarrow X$. (b) Our One2one CycleGAN. We propose to realize one-to-one mapping by learning ONLY one self-inverse function G for the mapping between two domains bidirectionally. It contains only one mapping function $G: X \leftrightarrow Y$.}
\label{fig:long}
\label{fig:onecol}
\end{figure}

The task of learning mappings between two domains from unpaired data has attracted a lot of attention, especially in the form of unpaired image-to-image translation \cite{CycleGAN2017,zhu2017toward,kim2017learning,liu2017unsupervised}. Thanks to the pioneer work of GAN\cite{goodfellow2014generative} and cycleGAN \cite{zhu2017unpaired}, recent works \cite{song2018geometry,pumarola2018ganimation,zhang2018translating,zhang2018task,choi2018stargan,huang2018multimodal,lee2018diverse,almahairi2018augmented,zhang2019harmonic,shen2019towards} have shown promising result for unpaired image-to-image translation. This task is very important because paired data are not available in many cases and the paired information are difficult or time-consuming to get. For example, in the medical field of cross domain medical image segmentation\cite{zhang2018task,zhang2018translating}: with the brain CT image semantic label and without the brain MRI semantic label, the goal is to generate semantic label for the brain MRI image.  Amazing works \cite{zhang2018task,zhang2018translating} like this cross domain image segmentation task in the medical image application could be further improved if unpaired image translation can be unique and more accurate. In many cases, the information source like patient is unique. For example, there is only a brain MRI image for a patient, but there should be a unique CT brain image for the same patient. This uniqueness requirement can be called one-to-one mapping of the brain CT and the brain MRI image from the same patient. If the unpaired image-to-image translation can achieve this one-to-one mapping, the cross-domain medical image segmenation performance can be further improved. However, existing method can not meet this requirement.

As mentioned above, the major limitation of existing methods for unpaired image-to-image translation like CycleGAN is that they can not realize one-to-one mapping which is necessary in many cases like the information source of the unpaired image is unique. Without the pairing information, CycleGAN using the distribution constraint allows many-to-many mappings. To reduce the space of possible mappings and improve in finding a more unique mapping, their models add an essential cycle-consistency constraint. The cycle-consistency constraint enforces a stronger connection cross domains by requiring the input image and the output image to be close. The output image is generated by first mapping from source domain to target domain then mapping back to the source domain. But this only reduces the many-to-many mapping to many-to-one mapping or one-to-many mapping cross domains, which will be illustrate in detail in section 3.2 and 3.3.

With the success of image generation\cite{goodfellow2014generative,radford2015unsupervised} model Generation Adversarial Networks(GANs) and unsupervised mapping methods like CycleGAN \cite{zhu2017unpaired}, motivated by the recent works \cite{shen2019towards2,jacobsen2018revnet} of exploring invertibility of convolutional neural networks (CNNs), we propose to learn a one-to-one mapping between domains from unpaired data to compensate the limitation of the exsiting methods such us CycleGAN.

Specifically, we enforce the generator of the CycleGAN as a self-inverse function to realize a one-to-one mapping. So we call our proposed method One2one CycleGAN. When a function $G$ is self-inverse, illustrated as
\begin{equation}
    G=G^{-1},
\end{equation} 
it guarantees a one-to-one mapping. We use the CycleGAN~\cite{zhu2017unpaired} as the baseline framework for image-to-image translation. To impose the self-inverse property, we implement {\it a simple idea} of augmenting the training samples by switching inputs and outputs during training. However, as we will demonstrate empirically, {\it this seemingly simple idea makes a genuinely big difference!}

The distinct feature of our self-inverse network is that it learns one network to perform both forward ($X \rightarrow Y$: from X to Y) and backward ($Y \rightarrow X$: from Y to X) translation tasks.
It contrasts with the state-of-the-art approaches which typically learn two separate networks, one for forwarding translation and the other for backward translation. As a result, it enjoys several benefits. First, it halves the necessary parameters, assuming that the self-inverse network and the two separate networks share the same network architecture. Second, it automatically doubles the sample size, a great feature for any data-driven models, thus becoming less likely to over-fit the model. 

One key question arises: Is it feasible to learn such a self-inverse network for image-to-image translation? We can not theoretically prove this existence; however, we experimentally demonstrate so. Intuitively, such an existence is related to the redundancy in the expressive power of the deep neural network. Even given a fixed network architecture, the function space for a network that translates an image from $A$ to $B$ is large enough, that is, there are many neural networks with different parameters capable of doing the same translation job. The same holds for the inversion network. Therefore, the overlap between these two spaces, in which the self-inverse network resides, does exist.

Our contribution are as follows: (i) We introduce the One2one CycleGAN model for learning one-to-one mappings cross domains in an unsupervised way. (ii) We show that our model can learn mappings that generate a more accurate output for each input. (iii) We evaluate our method in extensive experiments on a variety of datasets, including cross-modal medical image synthesis, object transfiguration, and semantic labeling, consistently demonstrate clear improvement over the CycleGAN method both qualitatively and quantitatively. Especially our proposed method reaches the state-of-the-art result on the cityscapes benchmark dataset for the label to photo unpaired directional image translation.

\begin{figure}[t]
\begin{center}
  \includegraphics[width=\linewidth]{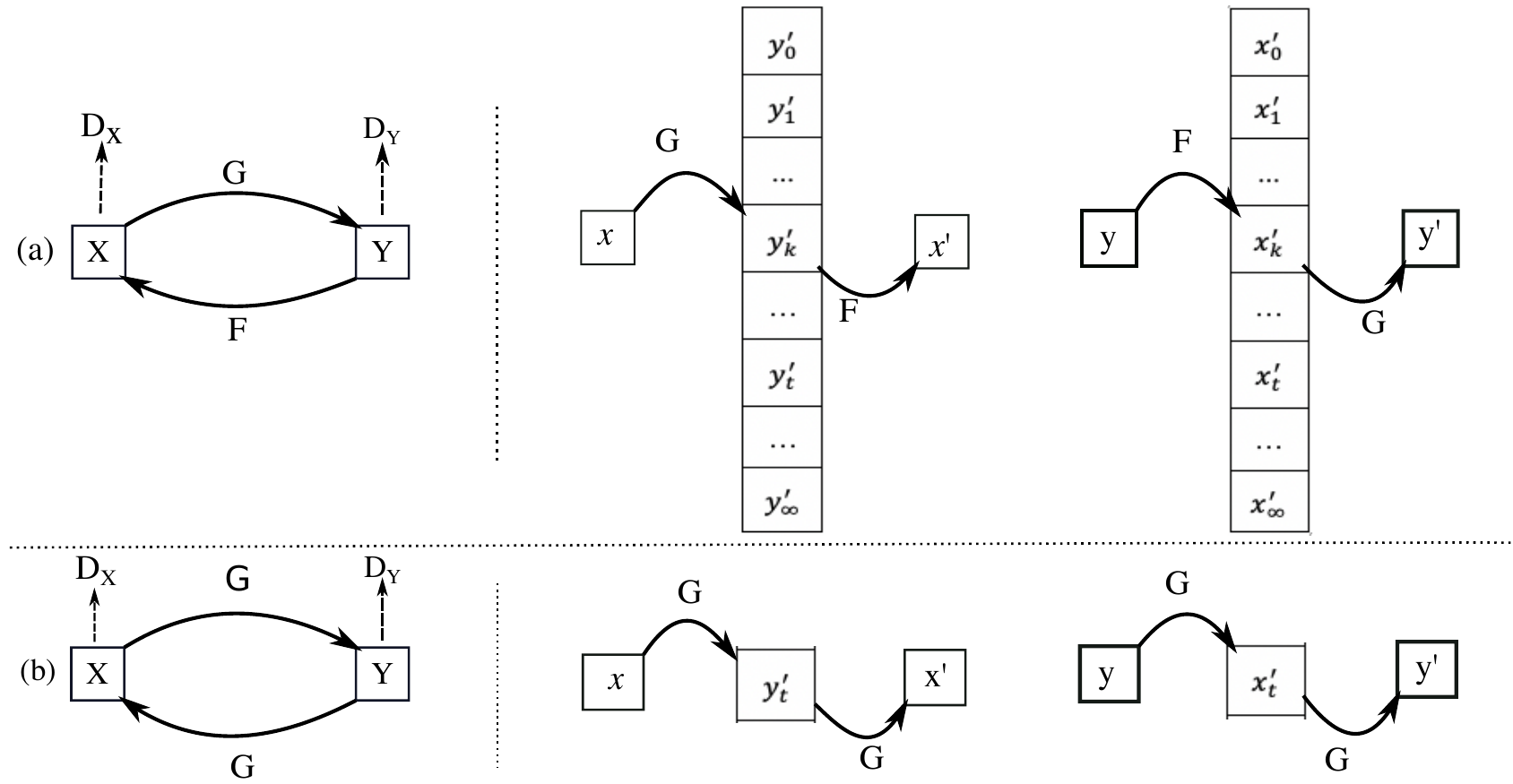}
\end{center}
   \caption{(a) The mapping routes of CycleGAN. The limitations of the CycleGAN model is that it allows biased and non-unique unpaired image translation. For the mapping route $x\rightarrow x'$, the mapping $G: x\rightarrow y'$ is a one to many mapping which result that x can be mapped to infinity possible $y'$, Let's denote the unique target is $y_t'$ and the actually mapped result is $y_k'$; The mapping $F: y_k'\rightarrow x'$ is a many to one mapping. As a result, there is allowable bias between the target $y_t'$ and the prediction $y_k'$. Similarly,for the mapping route $y\rightarrow y'$, the mapping $F: y\rightarrow x'$ is a one to many mapping which result that y can be mapped to infinity possible $x'$, Let's denote the unique target is $x_t'$ and the actually mapped result is $x_k'$; The mapping $G: x_k'\rightarrow y'$ is a many to one mapping. As a result, there is allowable bias between the target $x_t'$ and the prediction $x_k'$.  (b) The mapping routes of one2one CycleGAN. The motivation of one2one CycleGAN is to realize unique and accurate unpaired image translation. The mapping function G is self inverse function with the one-to-one mapping property. For the mapping route $x\rightarrow x'$, the mapping $G: x\rightarrow y'$ is a one to one mapping which result that x is only mapped to the unique target is $y_t'$. The mapping $F: y_t'\rightarrow x'$ is also a one to one mapping. As a result, there is no bias between the target and the prediction. Similarly, for the mapping route $y\rightarrow y'$, the mapping $F: y\rightarrow x'$ is a one to one mapping which result that y can only be mapped to the unique target is $x_t'$. The mapping $G: x_t'\rightarrow y'$ is also a one to one mapping. As a result, there is no bias between the target and the prediction.}
\label{fig:long}
\label{fig:onecol}
\end{figure}
\section{Literature Review}

Iosla et al.~\cite{isola2017image} presented the seminar work of image-to-image translation that offered a general-purpose solution, and Goodfellow et al. proposed to use the generative adversarial network (GAN)~\cite{goodfellow2014generative} for the first time in the literature. While paired data are assumed in~\cite{isola2017image}, later Zhu et al.~\cite{zhu2017unpaired} proposed the CycleGAN approach for addressing the unpaired setting using the so-called cyclic constraints. There are many recent advances that use guidance information~\cite{song2018geometry,pumarola2018ganimation}, impose different constraints~\cite{gan2017triangle,lu2018guiding,zhang19harmonic}, or deal with multiple domains\cite{zhu2017toward,choi2018stargan,huang2018multimodal,lee2018diverse}, etc. In this paper, we study unpaired image-to-image translation.

In addition to using the GAN that essentially enforces similarity in image distribution, other guidance information is used such as landmark points~\cite{song2018geometry}, contours~\cite{dekel2017smart}, sketches~\cite{lu2018image}, anatomical information \cite{pumarola2018ganimation} etc. In addition to cyclic constraint~\cite{zhu2017unpaired}, other constraints like ternary discriminative function~\cite{gan2017triangle}, optimal transport function~\cite{lu2018guiding}, smoothness over the sample graph~\cite{zhang19harmonic} are used as well.

Also, extensions were proposed to deal with video inputs~\cite{wang2018vid2vid,bansal2018recycle}, to synthesize images in high resolution~\cite{wang2018high}, to seek for diversity~\cite{mao2019mode}and to handle more than two image domains~\cite{zhu2017toward,choi2018stargan,huang2018multimodal,lee2018diverse}.
Furthermore, there are methods that leverage attention mechanism~\cite{ma2018gan,chen2018attention,mejjati2018unsupervised} and mask guidance~\cite{liang2018generative}. Finally, disentangling is a new emerging direction~\cite{huang2018multimodal,lee2018diverse}. 

In terms of works about inverse problem with neural networks,~\cite{jacobsen2018revnet} makes the CNN architecture invertible by providing an explicit inverse. Ardizzone et.al~\cite{ardizzone2018analyzing} prove the invertibility theoretically. More specifically, Kingma~\cite{kingma2018glow} shows the benefit of a invertible $1 \times 1$ convolution.

Different from a one-to-one mapping function are one-to-many, many-to-one, and many-to-many~\cite{almahairi2018augmented} \footnote{It is worth noting that recently there are quite some works focusing on addressing image-to-image translation among many domains, also the so-called one-to-many .} mapping functions. In \cite{isola2017image}, the well-studied scenarios of labels-to-scenes, edge-to-photo are more likely one-to-many mapping as it is possible that multiple photos (scenes) have the same edge (label) information. The colorization example is also one-to-many. From an information theory perspective, the entropy of the edge map (label) is low while that of the photo is high. When an image translation goes from an information-gaining direction, that is, from low-entropy to high-entropy, its mapping leans towards one-to-many. Similarly, if it goes from an information-losing direction, then its mapping leans toward many-to-one. If the information level of both domains is close (or information-similar), then the mapping is close to one-to-one.
In \cite{isola2017image}, the examples of Monet-to-photo, summer-to-winter are closer to one-to-one mapping as the underlying contents of both images before and after translation are regarded the same but the styles are different, which does not change the image entropy significantly. For image to image translation, many works has been done to diversify the output~\cite{almahairi2018augmented,liu2017unsupervised,lee2018diverse,huang2018multimodal,zhu2017toward,lee2019drit++}, while not too many work has been done to make the output unique~\cite{shen2019towards}. Our work goes to the latter direction.

Although there are so many research works on image-to-image translation, the perspective of learning a one-to-one mapping network has not been fully investigated, with the exception of~\cite{lu2018guiding}. In~\cite{lu2018guiding}, Lu et al. show that CycleGAN can not theoretically guarantee the one-to-one mapping property and propose to use an optimal transport mechanism to mitigate this issue. However, like GAN, the optimal transport method also measures the similarity in image distribution; hence the one-to-one issue is not fully resolved. By contrast, our self-inverse learning comes with a guarantee that the learned network realizes a one-to-one mapping. %To the best of our knowledge, the proposed method is the only one with such a guarantee.

\section{Unsupervised Learning of One-to-one Mappings between Domains}
\subsection{Problem setting}
For any two domains X and Y with only unpaired elements available, we assume there exists a mapping, potentially one-to-one mapping, between the elements of each domain. The goal to make sure there is a unique target element in the target domain to match a element in the source domain. The objective is to recover this mapping. Since there are only unpaired samples available, this goal is realized by matching the distributions $p_d(x)$ and $p_d(y)$ of each domain. This can be treated as a conditional generating task. The true conditionals $p(x|y)$ and $p(x|y)$ are estimated from the true marginals. To be able to uncover this mapping, the elements in both domain X and domain Y are highly dependent.
\subsection{CycleGAN model}

As shown in Figure 1, the CycleGAN model \cite{zhu2017unpaired} solves this problem by estimating these two conditionals with two separated mappings functions $G: X \rightarrow Y$ and $F: Y \rightarrow X$. Both of the mapping functions are parameterized with identical neural networks and constrained by the followings:

\begin{itemize}
    \item Distribution matching: The distribution of the each mapping output should match the distribution of the target domain. This constrain allows many-to-many mappings between the source domain X and the target domain Y and vice versa.
    \item Cycle-consistency: Each element is mapped from the source domain to the target domain, then mapped back to source domain. The output should be close to the input element. This constrains one-to-many mapping from source domain and many-to-one mapping from target domain to source domain.
\end{itemize}
\subsection{Limitations of CycleGAN for one-to-one mapping}
The main weakness of the CycleGAN model is that is can not realize one-to-one mapping for accurate and unique unpaired image translation. Based on the above constrains and the illustration in Figure 2, the CycleGAN model can not satisfy our problem's goal. Next, we show how to modify CycleGAN to meet the goal of our problem.

The distribution matching is implemented by GAN\cite{goodfellow2014generative}. The two mapping functions G and F implemented by neural networks are trained to fool the Discriminator $D_Y$ and $D_X$ respectively. The adversarial loss\cite{goodfellow2014generative} for mapping function G is
\begin{eqnarray}
 \mathcal{L}_{GAN}(G,D_X,X,Y) = \mathop{\mathbb{E}_{x \sim
 p_{data}(x)}}[\log D_Y(x)] \nonumber\\
 +\mathop{\mathbb{E}_{y\sim
 p_{data}(y)}}[\log(1-D_x(G(y)))].
\end{eqnarray}

The cycle consistency loss is:
\begin{eqnarray}
 \mathcal{L}_{cyc}(G,F) = \mathop{\mathbb{E}_{x \sim
 p_{data}(x)}}[||{F(G(x)) -x}||_1]\nonumber\\
 + \mathop{\mathbb{E}_{y \sim
 p_{data}(y)}}[||{G(F(y)) -y}||_1].
\end{eqnarray} 

The final objective for the mapping function G and F is
\begin{eqnarray}
\mathcal{L}(G,F,D_X,D_Y)= \mathcal{L}_{GAN}(G,D_Y,X,Y) \nonumber\\
+ \mathcal{L}_{GAN}(F,D_X,Y,X) + \lambda\mathcal{L}_{cyc}(G,F)
    \end{eqnarray}
and we aim to solves  
\begin{equation}
        (G^*,F^*) = arg \mathop{min}_{\mathbf{G,F}} \mathop{max}_{\mathbf{D_X,D_Y}}\mathcal{L}(G,F,D_X,D_Y).
\end{equation}
\section{Self-inverse Learning for Unpaired Image-to-image Translation}

In the section, we first show the property that the self-inverse function guarantees one-to-one (one2one) mapping. Then we discuss how to train a self-inverse CycleGAN network for image-to-image translation

\subsection{One-to-one property}

In image-to-image translation, we define a forward function as $Y=f_{X \rightarrow B}(X)$ that maps an image $X$ on domain $A$ to another image $Y$ on domain $B$ and, similarly, an inverse function as $X=f^{-1}_{B \rightarrow A}(Y)$. When there is no confusion, we will skip the subscript (e.g., $A \rightarrow B$).

\underline{Property}: If a function $Y=f(X)$ is self-inverse, that is $f=f^{-1}$, then the function $f$ defines a one-to-one mapping, that is, $Y_1 = Y_2$ if and only if $X_1=X_2$.

\underline{Proof:} 

[$\Rightarrow$] If $X_1 = X_2$, then $Y_1 = f(X_1) = f(X_2) =Y_2$.

[$\Leftarrow$] 
If $Y_1 = Y_2$, then $X_1 = f^{-1}(Y_1)= f^{-1}(Y_2)=X_2$ as long as the inverse function exists, which is the case for a self-inverse function as $f^{-1} = f$. $\#$

\subsection{One-to-one benefits}
There are several advantages in learning a self-inverse network to have the one-to-one mapping property.

(1) From the perspective of the application, only one self-inverse function can model both tasks $A$ and $B$ and it is a novel way for multi-task learning. As shown in Figure 1, the self-inverse network generates an output given an input, and vice versa, with only one CNN and without knowing the mapping direction. It is capable of doing both tasks within the same network, simultaneously. In comparison to separately assigning two CNNs for tasks $A$ and $B$, the self-inverse network halves the necessary parameters, assuming that the self-inverse network and the two CNNs share the same network architecture as shown in Figure 1.

(2) It automatically doubles the sample size, an important feature for any data-driven models, thus it is less likely to over-fit the model. The self-inverse function $f$ has the co-domain $ Z = X \cup Y$. If the sample size of either domain $X$ or $Y$ is $N$, then the sample size for domain $Z$ is $2N$. As a result, the sample size for both tasks $A$ and $B$ are doubled, becoming a novel method for data augmentation to mitigate the over-fitting problem.

(3) As shown in Figure 2, In the unpaired image-to-image translation setting, the goal is to minimize the distribution gap between the two domains. The state-of-art methods can realize this but can not guarantee an ordered mapping or bijection between the two domains. This results in variations for the generated images.

(4) The one-to-one mapping is a strict constraint. Therefore, forcing a CNN model as a self-inverse function can shrink the target function space.

\subsection{One-to-one CycleGAN}

We are inspired by the basic formulation of CycleGAN~\cite{zhu2017unpaired}. 
In CycleGAN, there are two generators $Y=F(X)$ and $X=G(Y)$, two discriminators $D_x$ and $D_y$, and one joint object function.
In our one2one CycleGAN, we have one shared generator $G$ and still two discriminators $D_x$ and $D_y$. Instead of having a joint objective for the dual-mappings, our proposed method has two separate objective functions, one for each of two mapping directions.

\subsubsection{Separated loss functions}
Compared to CycleGAN that uses a joint loss for both image transfer directions, our method have two separate losses, one for each image transfer direction.
%\textbf{Adversarial Loss}
For the mapping function $G: X \rightarrow Y$ and its discriminator $D_Y$, the adversarial loss is
\begin{eqnarray}
 \mathcal{L}_{GAN}(G,D_Y,X,Y) = \mathop{\mathbb{E}_{y \sim
 p_{data}(y)}}[\log D_Y(y)] \nonumber\\
 +\mathop{\mathbb{E}_{x\sim
 p_{data}(x)}}[\log(1-D_Y(G(x)))].
\end{eqnarray}
The cycle consistency loss is
\begin{equation}
 \mathcal{L}_{cyc}^x(G) = \mathop{\mathbb{E}_{x\sim
 p_{data}(x)}}[||{G(G(x)) -x}||_1].
\end{equation}

For the mapping function $G: Y \rightarrow X$ and its discriminator $D_X$, the adversarial loss is:
\begin{eqnarray}
 \mathcal{L}_{GAN}(G,D_X,X,Y) = \mathop{\mathbb{E}_{x \sim
 p_{data}(x)}}[\log D_Y(x)] \nonumber\\
 +\mathop{\mathbb{E}_{y\sim
 p_{data}(y)}}[\log(1-D_x(G(y)))].
\end{eqnarray}
The cycle consistency loss is:
\begin{equation}
 \mathcal{L}_{cyc}^y(G) = \mathop{\mathbb{E}_{y \sim
 p_{data}(y)}}[||{G(G(y)) -y}||_1].
\end{equation}

So, the final objective for the mapping function $X \rightarrow Y$ is
\begin{equation}
\mathcal{L}(G,D_Y)= \mathcal{L}_{GAN}(G,D_Y,X,Y) + \lambda_x\mathcal{L}_{cyc}^x(G),
\end{equation}
and the minimax optimization solves
\begin{equation}
        (G^*,D_Y^*)= arg \mathop{min}_{\mathbf{G}} \mathop{max}_{\mathbf{D_Y}}\mathcal{L}(G,D_Y).
\end{equation}

Similarly, the final objective for the mapping function $Y \rightarrow X$ is
\begin{equation}
\mathcal{L}(G,D_X)= \mathcal{L}_{GAN}(G,D_X,X,Y) + \lambda_y\mathcal{L}_{cyc}^y(G),
    \end{equation}
and the minimax optimization solves  
\begin{equation}
        (G^*,D_X^*) = arg \mathop{min}_{\mathbf{G}} \mathop{max}_{\mathbf{D_X}}\mathcal{L}(G,D_X).
\end{equation}

\begin{figure}[t]
\begin{center}
  \includegraphics[width=\linewidth]{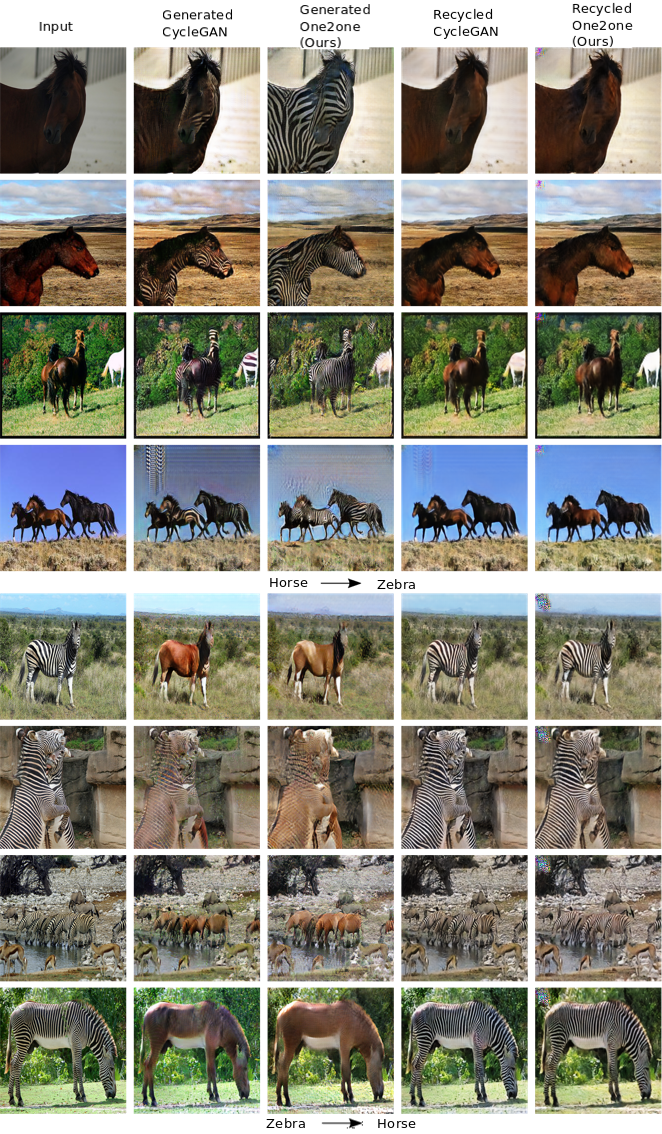}
\end{center}
   \caption{Visual comparison for horse$\leftrightarrow$zebra.}
\label{fig:long}
\label{fig:onecol}
\end{figure}

\subsection{Self-inverse implementation}

We apply the proposed method based on the framework of CycleGAN \cite{zhu2017unpaired}. To have a fair comparison with CycleGAN, we adopt the architecture of (Johnson et al., 2016) as the generator and the
PatchGAN \cite{isola2017image} as the discriminator. The log likelihood objective in the original GAN
is replaced with a least-squared loss \cite{johnson2016perceptual} for more stable training. We resize the input
images to $256 \times 256$.  The loss weights are set as $\lambda_x = \lambda_y = 10$. Following CycleGAN,
we adopt the Adam optimizer ~\cite{kingma2014adam} with a learning rate of 0.0002. Similarly, we use a pool size of 50. The learning
rate is fixed for the first 100 epochs and linearly decayed to zero over the next 100 epochs on Yosemite and apple2orange datasets. The learning
rate is fixed for the first 4 epochs and linearly decayed to zero over the next 3 epochs on the BRATS dataset. The learning
rate is fixed for the first 90 epochs and linearly decayed to zero over the next 30 epochs on the Cityscapes dataset.

\subsection{Training details and optimization}
In our experiments, we use a batch size of 1. At each iteration, we randomly sample a batch of pair $(x_{i},y_{i})$, where samples  $\left \{x_{i}  \right \}_{i=1}^N \in X $ and $\left \{y_{i}  \right \}_{i=1}^M \in Y $. At any iteration $j$, we perform the following three steps:
\begin{itemize}
    \item Firstly, we feed $x_i$ as the input and $y_i$ as the target, then forward $G$ and back-propagate $G$; 
    \item Secondly, we feed $y_i$ as the input and $x_i$ as the target, then forward $G$ and back-propagate $G$; 
    \item Finally, we back-propagate $D_Y$ and $D_X$ individually.
\end{itemize}

\begin{table*}
\begin{center}
\begin{tabular}{|c|c|c|c|c|c|c|l|}
\hline
         &
         \multicolumn{3}{c}{Label $\rightarrow$ Photo} & \multicolumn{3}{c}{Photo $\rightarrow$ Label} \\
         \hline\hline
    Method&Pixel Acc.$\uparrow$  & Class Acc. $\uparrow$ & Class IoU $\uparrow$ &Pixel Acc.$\uparrow$  & Class Acc. $\uparrow$ & Class IoU $\uparrow$ \\
    CycleGAN  & 52.7 & 15.2 &11.0& \bf57.2&\bf21.0&\bf15.7\\
    DiscoGAN &45.0 & 11.1 & 7.0&45.2&10.9&6.3\\ 
    DistanceGAN & 48.5 & 10.9 &7.3&20.5&8.2&3.4\\
    UNIT & 48.5 & 12.9 &7.9&56.0&20.5&14.3\\
    \hline
    One2one CycleGAN (ours) & \bf{58.2} & \bf{18.9} & \bf14.3& 52.7 & 18.1& 13.0\\
    \hline
\end{tabular}
\end{center}
\caption{Results of Photo $\leftrightarrow$ Label translation on the Cityscapes dataset.}
\end{table*}

\section{Experiments}

In order to test the effect of the proposed method, we evaluate it on an array of applications: cross-modal medical image synthesis, object transfiguration, and style transfer.
Also we compare against several unpaired image-to-image translation methods: CycleGAN~\cite{zhu2017unpaired}, DiscoGAN~\cite{kim2017learning}, DistanceGAN~\cite{benaim2017one}, and UNIT~\cite{liu2017unsupervised}. We conduct a user study when the ground truth images are unknown and perform quantitative evaluation when the ground truth images are present.

\subsection{Datasets and results}
\textbf{Object transfiguration.}
we test our method on the horse $\leftrightarrow$ zebra task used in CycleGAN paper~\cite{zhu2017unpaired} with 2401 training images (939 horses and 1177 zebras) and 260 test images (120 horses and 140 zebras). This task has no ground truth for generated images and hence no quantitative evaluation is feasible. So we provide the qualitative results obtained in a user study. In the user study, we ask a user to rate his/her preferred image out of three randomly positioned images, one obtains from CycleGAN, one from DistanceGAN, and the other from one2one CycleGAN. Figure 4 shows examples of input and synthesized images and Table 1 summarize the use study results.

Figure 4 tells that one2one CycleGAN likely generates better quality images in an unsupervised fashion, especially in terms of the quality of zebra synthesis from the horse (refer to the first four rows). Our method generated more real and complete zebra content. From Table 1, it is clear that our one2one CycleGAN is the most favorable with a 75\% (77\%) preference percentage for the horse2zebra (zebra2horse) mapping direction. and DistanceGAN is the least favorable.  

we test our method on the apple $\leftrightarrow$ orange task~\cite{zhu2017unpaired} with 2014 training images (995 apples and 1019 orange) and 514 test images (248 apples and 266 oranges). This task has no ground truth for generated images and hence no quantitative evaluation is feasible. Figure 5 shows examples of input and synthesized images. There are failure cases in rows 1,2,4 from CycleGAN while our model generates normal images.

\begin{figure}[t]
\begin{center}
  \includegraphics[width=\linewidth]{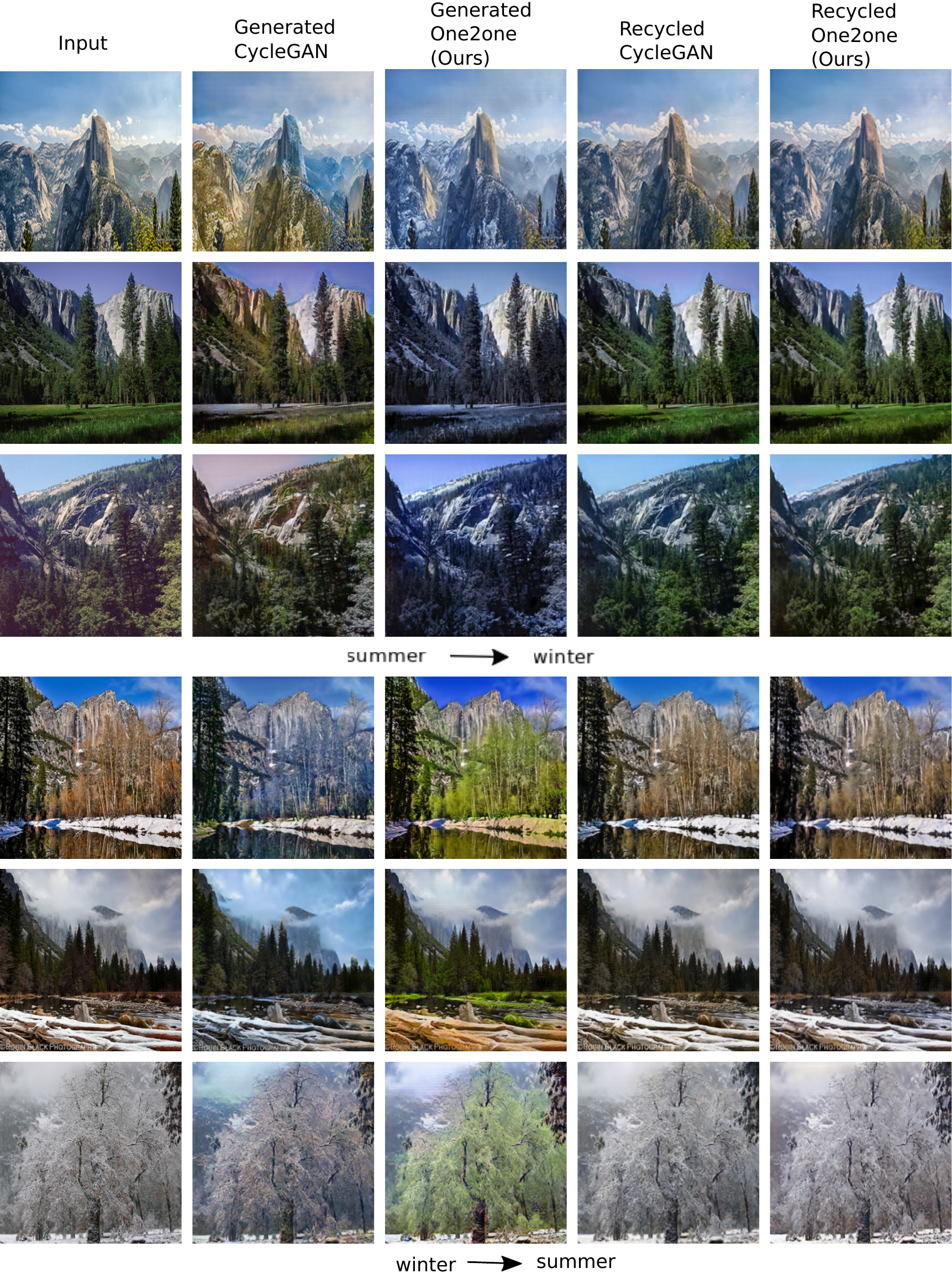}
\end{center}
   \caption{Visual comparison for summer$\leftrightarrow$winter on yosemite.}
\label{fig:long}
\label{fig:onecol}
\end{figure}

\begin{table}
\begin{center}
 \begin{tabular}{|c|c|c|c|l|}
 \hline
    Direction&Metric&Cycle & Distance &One2one \\
    \hline\hline
    horse2zebra & Prefer pct. $\uparrow$ &25\% & 0 & 75\%\\
    zebra2horse & Prefer pct. $\uparrow$ &23\%& 0 & 77\%\\
    \hline
\end{tabular}
\end{center}
\caption{Results of user study on the horse to zebra dataset.}
\end{table}

\begin{figure}[t]
\begin{center}
  \includegraphics[width=\linewidth]{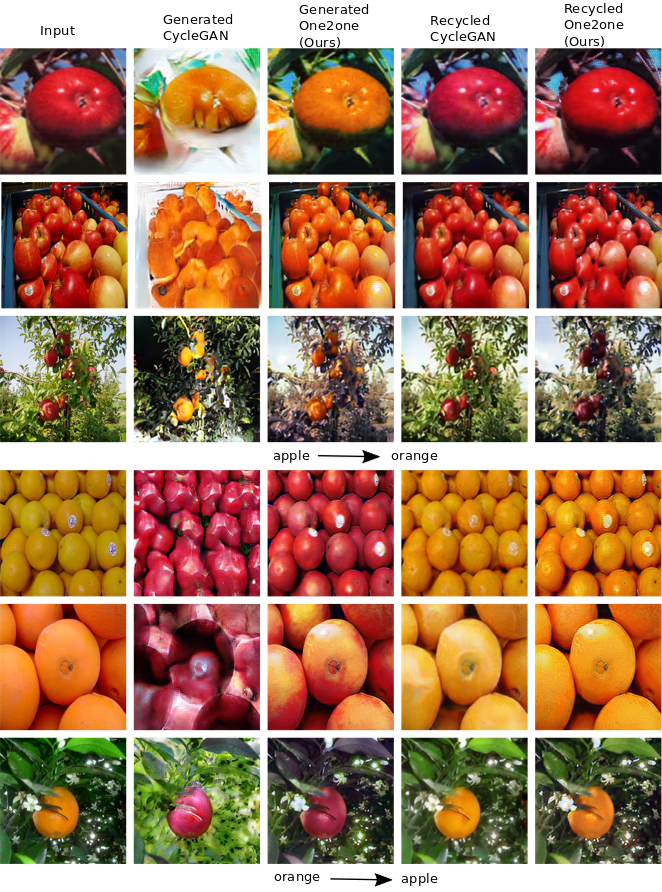}
\end{center}
   \caption{Visual comparison for apple$\leftrightarrow$orange.}
\label{fig:long}
\label{fig:onecol}
\end{figure}

\begin{figure*}[t]
\begin{center}
  \includegraphics[width=\linewidth]{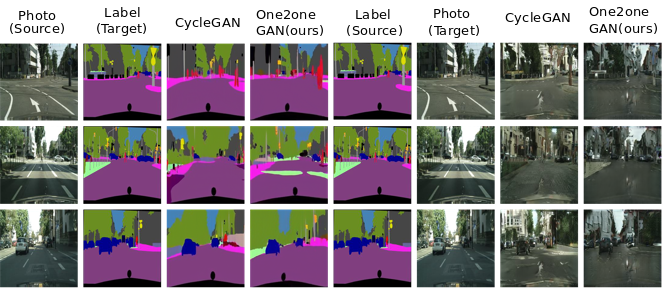}
\end{center}
   \caption{Visual comparison for photo$\leftrightarrow$label on the Cityscapes.}
\label{fig:long}
\label{fig:onecol}
\end{figure*}

\textbf{Cross-modal medical image synthesis.}
This task evaluates cross-modal medical image synthesis. The
models are trained on the BRATS dataset \cite{menze2015multimodal} which contains paired MRI data to
allow quantitative evaluation. It contains ample multi-institutional routine clinically-acquired pre-operative multi modal MRI scans of glioblastoma (GBM/HGG) and lower grade glioma (LGG) images. There are 285 3D volumes for training and 66 3D volume for the test. The $T_1$ and $T_2$ images are selected for our bi-directional image synthesis. All the 3D volumes are preprocessed to one channel image of size 256 x 256 x 1.
We use the Peak Signal-to-Noise Ratio (PSNR) and Structural Similarity Index Measure (SSIM) to
evaluate the quality of generated images.

\begin{figure*}[t]
\begin{center}
  \includegraphics[width=\linewidth]{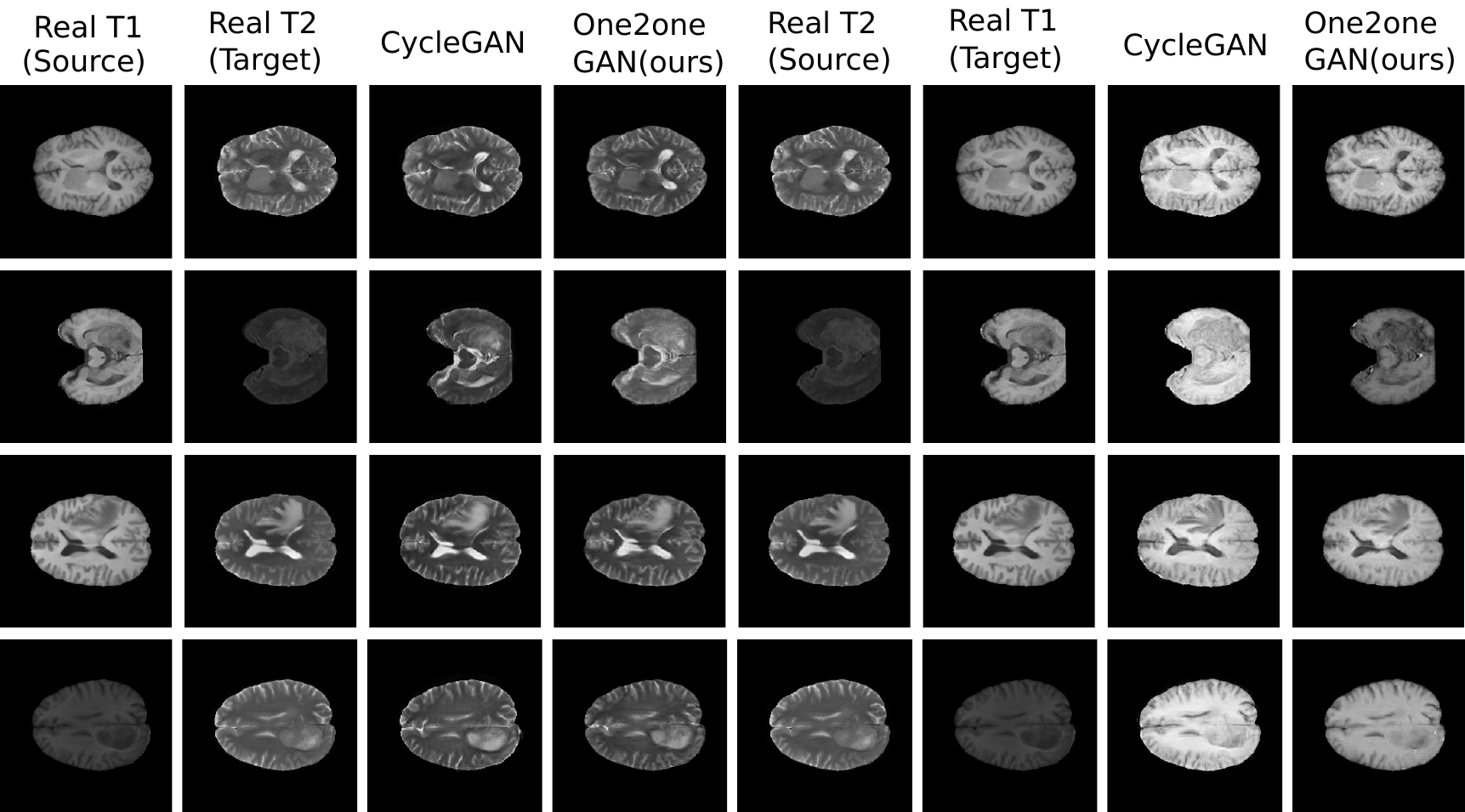}
\end{center}
   \caption{Qualitative comparison for T1$\leftrightarrow$T2 on BRATS datasets.}
\label{fig:long}
\label{fig:onecol}
\end{figure*}

As shown in Table 2, on the $T_1 \rightarrow T_2$ image synthesis direction, our one2one model outperforms the CycleGAN model on PSNR by 6.0\%. The qualitative result is shown in columns 3 and 4 in Figure 7. On the $T_2 \rightarrow T_1$ image synthesis direction, our one2one model outperforms the CycleGAN model on PSNR by 5.0\%.  The qualitative result is shown in columns 7 and 8 in Figure 7.

\begin{table}
\begin{center}
  \begin{tabular}{|c|c|c|l|}
    \hline
    Direction&Method&PSNR $\uparrow$ & SSIM $\uparrow$ \\
    \hline\hline
    T1 $\rightarrow$ T2 & CycleGAN & 20.79 &0.85\\
    T1 $\rightarrow$ T2 & One2one CycleGAN & \bf{22.03} & \bf{0.86}\\ \hline
    T2 $\rightarrow$ T1 & CycleGAN & 17.47 &0.81\\
    T2 $\rightarrow$ T1 & One2one CycleGAN & \bf{18.31} & \bf{0.82}\\
    \hline
\end{tabular}
\end{center}
\caption{Evaluation of cross-modal medical image synthesis on the BRATS datase.}
\end{table}

\textbf{Semantic labeling.}
We also test our method on the labels $\leftrightarrow$ photos task using the Cityscapes
dataset \cite{cordts2016cityscapes} under the unpaired setting as in the original CycleGAN paper. For quantitative evaluation, in line with previous work, for labels $\rightarrow$ photos we adopt the ``FCN score" \cite{isola2017image}, which evaluates how interpretable the generated photos are according to a semantic
segmentation algorithm. For photos $\leftarrow$ labels, we use the standard segmentation metrics, including
per-pixel accuracy, per-class accuracy, and mean class Intersection-Over-Union (Class IoU).
The quantitative result is shown in Table 3. Our model reaches the state-of-the-art on the label $\rightarrow$ photo direction image synthesis under this unpaired setting. The pixel accuracy outperforms the second best result by 10.4 \%; The class accuracy outperforms the second best result by 24.3 \%; The class IoU outperforms the second best result by 30.0 \%. On the photo $\rightarrow$ label direction, our model reaches comparable results.

\begin{table}
\begin{center}
  \begin{tabular}{|c|c|c|c|l}
  \hline
    Direction&Metric&Cycle  &One2one \\
    \hline\hline
    summer2winter & Prefer pct. $\uparrow$ &34\%  & 66\%\\
    winter2summer & Prefer pct. $\uparrow$ &41\% & 59\%\\
    \hline
\end{tabular}
\end{center}
\caption{Results of user study on the summer to winter Yosemite dataset.}
\end{table}

The qualitative result is shown in Figure 6. Compared with CycleGAN which is the second best result in the label $\rightarrow$ photo direction, our model has clearly better visual results.
On the photo $\rightarrow$ label direction, our model also have a comparable or better result.

\textbf{Style Transfer.}
We also test our method on the summer $\leftrightarrow$ winter style transfer task using the Yosemite
dataset under the unpaired setting as in the original CycleGAN paper. As shown in Figure 4 for the qualitative result, our method has better visual result in both directions of style transfer. We also do a similar user study by providing the generated image from the test set by our model and the CyecleGAN to users. The result is in Table 4. The user study results show that our model has a higher preference than CycleGAN.

\section{Conclusions}

We have presented an approach for enforcing the learning of a one-to-one mapping function for unpaired image-to-image translation. The proposed one-to-one CycleGAN consistently outperforms the baseline CycleGAN model and other state-of-the-art unsupervised approaches in terms of various qualitative and quantitative metrics.

\section{Acknowledgment}
The work was supported by grants from Siemens Healthineers (Siemens 082387).
{\small
\bibliographystyle{ieee}
\bibliography{egpaper_final}
}

\end{document}